\documentclass[10pt,letterpaper,twocolumn]{article}
\providecommand{\tabularnewline}{\\}

\usepackage{cvpr}
\usepackage{times}
\usepackage{epsfig}
\usepackage{graphicx}
\usepackage{amsmath}
\usepackage{amssymb}

\cvprfinalcopy 

\ifcvprfinal \fi

\usepackage[breaklinks=true,bookmarks=false]{hyperref}

\usepackage{array}
\usepackage{textcomp}
\usepackage{multirow}
\usepackage{setspace}

\usepackage{amsthm}
\theoremstyle{definition}
\newtheorem{definition}{Definition}[section]

\begin{document}

\title{Semi-supervised Vocabulary-informed Learning }

\author{Yanwei Fu and Leonid Sigal\\
Disney Research  \\
{\tt\small y.fu@qmul.ac.uk, lsigal@disneyresearch.com}
}

\maketitle
\begin{abstract}
Despite significant progress in object categorization, in recent years,
a number of important challenges remain; mainly, ability to learn
from limited labeled data and ability to recognize object classes
within large, potentially open, set of labels. Zero-shot learning
is one way of addressing these challenges, but it has only been shown
to work with limited sized class vocabularies and typically requires
separation between supervised and unsupervised classes, allowing former
to inform the latter but not vice versa. We propose the notion of
semi-supervised vocabulary-informed learning to alleviate the above
mentioned challenges and address problems of supervised, zero-shot
and open set recognition using a unified framework. Specifically,
we propose a maximum margin framework for semantic manifold-based
recognition that incorporates distance constraints from (both supervised
and unsupervised) vocabulary atoms, ensuring that labeled samples
are projected closest to their correct prototypes, in the
embedding space, than to others. We show that resulting model shows
improvements in supervised, zero-shot, and large open set recognition,
with up to 310K class vocabulary on AwA and ImageNet datasets.

\end{abstract}
\vspace{-0.2in}

\section{Introduction} \thispagestyle{empty}

Object recognition, and more specifically object categorization, has
seen unprecedented advances in recent years with development of convolutional
neural networks (CNNs) \cite{KrizhevskySH12}. However, most successful
recognition models, to date, are formulated as supervised learning
problems, in many cases requiring hundreds, if not thousands, labeled
instances to learn a given concept class \cite{deng2009imagenet}.
This exuberant need for large labeled datasets has limited recognition
models to domains with 100's to few 1000's of classes. Humans, on the other
hand, are able to distinguish beyond $30,000$ basic level categories
\cite{object_cat_1987}. What is more impressive, is the fact that
humans can learn from few examples, by effectively leveraging information
from other object category classes, and even recognize objects without
ever seeing them (\eg, by reading about them on the Internet). This
ability has spawned research in few-shot and zero-shot learning.

\begin{figure}
\begin{centering}
\includegraphics[width=0.97\columnwidth]{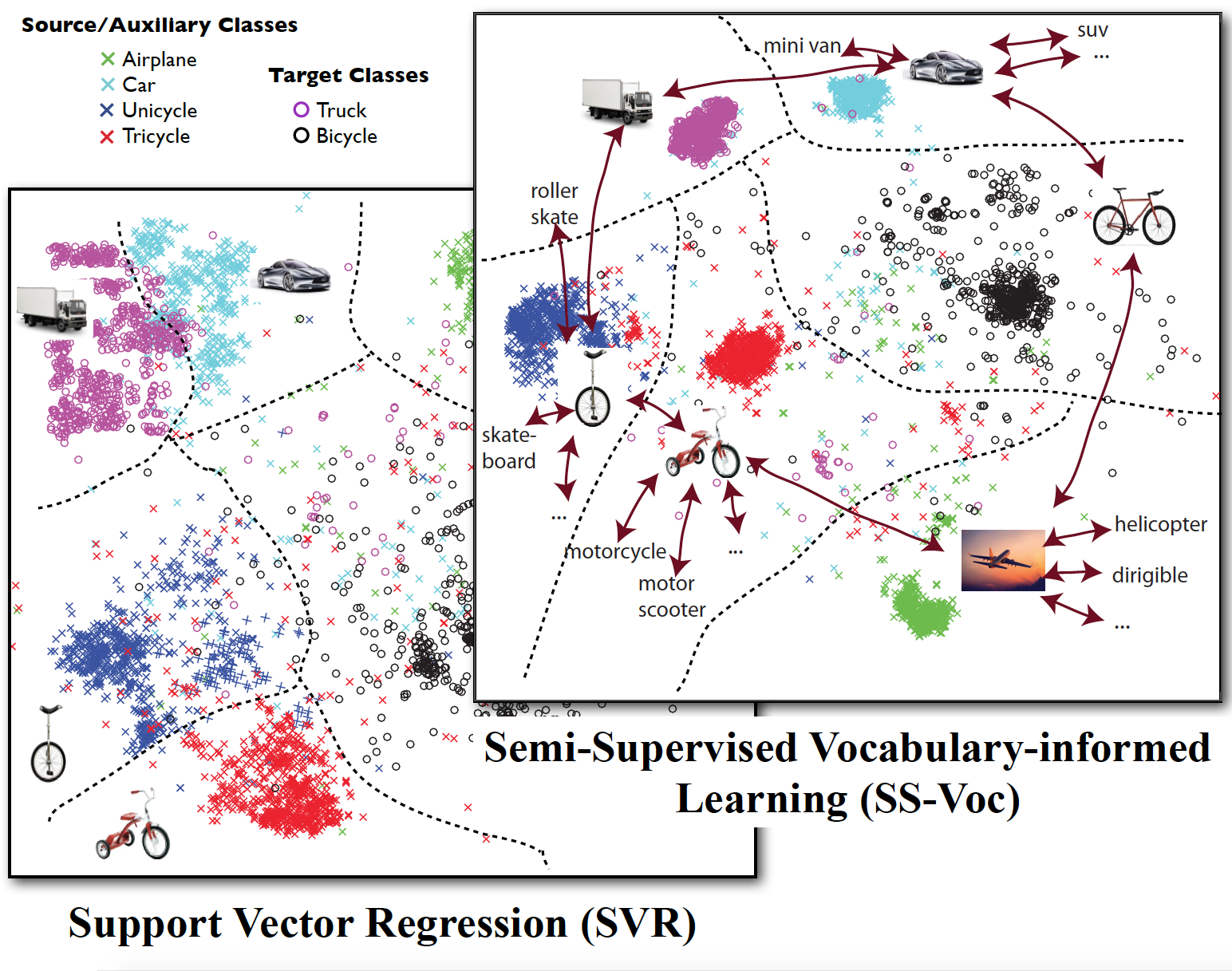}
\par\end{centering}
\protect\protect\caption{\label{fig:intro} {\bf Illustration of the semantic embeddings} learned (left)
using support vector regression ({\bf SVR}) and (right) using the proposed semi-supervised vocabulary-informed
({\bf SS-Voc}) approach. In both cases, t-SNE visualization is used to illustrate samples from $4$ 
source/auxiliary classes (denoted by $\times$) and $2$ target/zero-shot classed (denoted by $\circ$)
from the ImageNet dataset. Decision boundaries, illustrated
by dashed lines, are drawn by hand for visualization. 
Note, that (i) large margin constraints in SS-Voc, both among the source/target classes and the 
external vocabulary atoms (denoted by arrows and words), and (ii) fine-tuning of the semantic word space, 
lead to a better embedding with more compact and separated classes (\eg, see  {\em truck} and {\em car} or 
{\em unicycle} and {\em tricycle}). } 
\vspace{-0.25in}
\end{figure}

Zero-shot learning (ZSL) has now been widely studied in a variety
of research areas including neural decoding by fMRI images \cite{palatucci2009zero_shot},
character recognition~\cite{zero_data_AAAI2008}, face verification
\cite{kumar2009}, object recognition \cite{lampert13AwAPAMI}, and
video understanding~\cite{yanweiPAMIlatentattrib,zuxuan_2016_CVPR}. Typically, zero-shot 
learning approaches aim to recognize instances from the unseen
or unknown testing {\em target} categories by transferring information,
through intermediate-level semantic representations, from known observed
{\em source} (or auxiliary) categories for which many labeled instances
exist. In other words, supervised classes/instances, are used as context
for recognition of classes that contain no visual instances at training
time, but that can be put in some correspondence with supervised classes/instances.
As such, a general experimental setting of ZSL is that the classes
in target and source (auxiliary) %
dataset are disjoint and typically the learning is done on the source
dataset and then information is transferred to the target dataset,
with performance measured on the latter.

This setting has a few important drawbacks: (1) it assumes that target
classes cannot be mis-classified as source classes and vice versa;
this greatly and unrealistically simplifies the problem; (2) the target
label set is often relatively small, between ten \cite{lampert13AwAPAMI}
and several thousand unknown labels \cite{DeviseNIPS13}, compared
to at least $30,000$ entry level categories that humans can distinguish;
(3) large amounts of data in the source (auxiliary) classes are required,
which is problematic as it has been shown that most object classes
have only few instances (long-tailed distribution of objects in the
world \cite{torralba80M}); and (4) the vast open set
vocabulary from semantic knowledge, defined as part of ZSL~\cite{palatucci2009zero_shot},
is not leveraged in any way to inform the learning or source class recognition.

A few works recently looked at resolving (1) through class-incremental
learning \cite{Scheirer_2013_TPAMI,RichardNIPS13} which
is designed to distinguish between seen (source) and unseen (target)
classes at the testing time and apply an appropriate model -- supervised
for the former and ZSL for the latter. However, (2)--(4) remain largely
unresolved.  
In particular, while (2) and (3) are artifacts of the
ZSL setting, (4) is more fundamental. For example, consider learning
about a {\em car} by looking at image instances in Fig.\ref{fig:intro}.
Not knowing that other motor vehicles exist in the world, one may
be tempted to call anything that has 4-wheels a {\em car}. 
As a result the zero-shot class {\em truck}
may have large overlap with the {\em car} class (see Fig.\ref{fig:intro} [SVR]). 
However, imagine knowing that there also exist many other motor vehicles 
(trucks, mini-vans, {\em etc}). Even without having visually seen
such objects, the very basic knowledge that they {\em exist} in
the world and are closely related to a {\em car} should, in principal,
alter the criterion for recognizing instance as a {\em car} (making the recognition criterion stricter in this case). 
Encoding this in our [SS-Voc] model results in better separation among classes.

To tackle the limitations of ZSL and towards the goal of generic open
set recognition, we propose the idea of \textit{semi-supervised vocabulary-informed
learning}. Specifically, assuming we have few labeled training instances
and a large open set vocabulary/semantic dictionary (along with textual
sources from which statistical semantic relations 
among vocabulary atoms can
be learned), the task of semi-supervised vocabulary-informed learning
is to learn a model that utilizes semantic dictionary to help train
better classifiers for observed (source) classes and unobserved
(target) classes in supervised, zero-shot and open set image recognition
settings. Different from standard semi-supervised learning, we do
not assume unlabeled data is available, to help train classifier,
and only {\em vocabulary} over the target classes is known.

\vspace{0.1in}
\noindent
\textbf{Contributions:} Our main contribution is to propose
a novel paradigm for potentially open set image recognition: \textit{semi-supervised
vocabulary-informed learning} (\textbf{SS-Voc}), which is capable of utilizing vocabulary
over unsupervised items, during training, to improve recognition. A unified maximum
margin framework is used to encode this idea in practice.  Particularly,
classification is done through nearest-neighbor distance to class
prototypes in the semantic embedding space, and we encode a set of
constraints ensuring that labeled images project into semantic space
such that they end up closer to the correct class prototypes than
to incorrect ones (whether those prototypes are part of the source
or target classes). We show that word embedding (word2vec) can be
used effectively to initialize the semantic space. Experimentally,
we illustrate that through this paradigm: we can achieve competitive
supervised (on source classes) and ZSL (on target classes) performance,
as well as open set image recognition performance with large number
of unobserved vocabulary entities (up to $300,000$); effective learning
with few samples is also illustrated. 

\noindent \vspace{-0.05in}

\section{Related Work}

\vspace{-0.05in}

\noindent 
\textbf{One-shot Learning:} While most of machine learning-based
object recognition algorithms require large amount of training data,
one-shot learning \cite{feifei2006one_shot} aims to learn object
classifiers from one, or only few examples. To compensate for the
lack of training instances and enable one-shot learning, {\em knowledge}
much be transferred from other sources, for example, by sharing features
\cite{bart2005cross_gen}, semantic attributes \cite{yanweiPAMIlatentattrib,lampert13AwAPAMI,transferlearningNIPS,rohrbach2010semantic_transfer},
or contextual information \cite{one_shot_TL_contexutal}. However,
none of previous works had used the open set vocabulary to help learn
the object classifiers.

\vspace{0.05in}
\noindent
\textbf{Zero-shot Learning:} ZSL aims to recognize novel classes
with no training instance by transferring {\em knowledge} from
source classes. ZSL was first explored with use of attribute-based
semantic representations~\cite{farhadi2009attrib_describe,fu2012attribsocial,yanweiPAMIlatentattrib,transductiveEmbeddingJournal,kumar2009,parikh2011relativeattrib}.
This required pre-defined attribute vector prototypes for each class,
which is costly for a large-scale dataset. Recently, semantic word
vectors were proposed as a way to embed any class name without 
human annotation effort;
they can therefore serve as an alternative semantic representation
\cite{embedding_akata,DeviseNIPS13,semantic_graph,ZSL_convex_optimization}
for ZSL. Semantic word vectors are learned from large-scale text corpus
by language models, such as word2vec \cite{distributedword2vec2013NIPS},
or GloVec \cite{GloVec}. However, most of previous work only use
word vectors as semantic representations in ZSL setting, but have
neither (1) utilized semantic word vectors explicitly for learning
better classifiers; nor (2) for extending ZSL setting towards open
set image recognition. A notable exception is \cite{ZSL_convex_optimization}
which aims to recognize 21K zero-shot classes given a modest vocabulary of 
1K source classes; we explore vocabularies that are up to an order of the 
magnitude larger -- 310K.  

\vspace{0.05in}
\noindent 
\textbf{Open-set Recognition:} The term ``open set recognition''
was initially defined in \cite{Scheirer_2014_TPAMIb,Scheirer_2013_TPAMI} and formalized in \cite{Bendale_2015_CVPR,Sattar_2015_CVPR}
which mainly aims at identifying whether an image belongs to a seen
or unseen classes. It is also known as 
class-incremental learning. However, none of them can further identify classes for  unseen instances. 
An exception is \cite{ZSL_convex_optimization} which 
augments zero-shot (unseen) class labels with source (seen) labels 
in some of their experimental settings. 
Similarly, we define the \emph{open set image recognition} as the
problems of recognizing the class name of an image from a potentially
very large open set vocabulary (including, but not limited to source and
target labels). Note that methods like \cite{Scheirer_2014_TPAMIb,Scheirer_2013_TPAMI}
are orthogonal but potentially useful here
-- it is still worth identifying seen or unseen instances to be recognized
with different label sets as shown in experiments. Conceptually
similar, but different in formulation and task, open-vocabulary object
retrieval~\cite{Guadarrama14:OOR} focused on retrieving objects
using natural language open-vocabulary queries. 

\vspace{0.05in}
\noindent 
\textbf{Visual-semantic Embedding:} Mapping between visual features
and semantic entities has been explored in two ways: (1) directly
learning the embedding by regressing from visual features to the semantic
space using Support Vector Regressors (SVR) \cite{farhadi2009attrib_describe,lampert13AwAPAMI}
or neural network \cite{RichardNIPS13}; (2) projecting visual features
and semantic entities into a common {\em new} space, such as SJE
\cite{embedding_akata}, WSABIE \cite{Weston:2011:WSU:2283696.2283856},
ALE \cite{labelembeddingcvpr13}, DeViSE \cite{DeviseNIPS13}, and
CCA \cite{yanweiembedding,transductiveEmbeddingJournal}. In contrast, our model trains
a better visual-semantic embedding from only few training instances
with the help of large amount of open set vocabulary items (using a 
maximum margin strategy). Our formulation is inspired by the
unified semantic embedding model of \cite{unified_model}, however,
unlike \cite{unified_model}, our formulation is built on word vector
representation, contains a data term, and incorporates constraints
to unlabeled vocabulary prototypes. 

\section{Vocabulary-informed Learning}
\vspace{-0.02in}
%
%
%
Assume a labeled source dataset $\mathcal{D}_{s}=\{\mathbf{x}_{i},z_{i}\}_{i=1}^{N_{s}}$
of $N_{s}$ samples, where $\mathbf{x}_{i}\in\mathbb{R}^{p}$ is the
image feature representation of image $i$ and $z_{i}\in\mathcal{W}_{s}$
is a class label taken from a set of English words or phrases $\mathcal{W}$; consequently,
$|\mathcal{W}_{s}|$ is the number of source classes. Further, suppose
another set of class labels for target classes $\mathcal{W}_{t}$,
such that $\mathcal{W}_{s}\cap\mathcal{W}_{t}=\emptyset$,
for which no labeled samples are available. We note that potentially
$|\mathcal{W}_{t}|>>|\mathcal{W}_{s}|$. Given a new test image feature
vector $\mathbf{x}^{*}$ the goal is then to learn a function $z^{*}=f(\mathbf{x}^{*})$,
using all available information, that predicts a class label $z^{*}$.
Note that the form of the problem changes drastically depending on
which label set assumed for $z^{*}$: Supervised learning: $z^{*}\in\mathcal{W}_{s}$;
Zero-shot learning: $z^{*}\in\mathcal{W}_{t}$; Open set recognition:
$z^{*}\in \{\mathcal{W}_{s},\mathcal{W}_{t}\}$ or, more generally, $z^{*}\in\mathcal{W}$. We posit
that a single unified $f(\mathbf{x}^{*})$ can be learned for all
three cases. 
We formalize the definition of semi-supervised vocabulary-informed learning (SS-Voc) as follows: 

\theoremstyle{definition}
\begin{definition}{ \em Semi-supervised Vocabulary-informed Learning (SS-Voc): } 
is a learning setting that makes use of complete vocabulary data ($\mathcal{W}$) during training. Unlike a more traditional ZSL that typically makes use of the vocabulary (\eg, semantic embedding) at test time, SS-Voc utilizes exactly the same data during training. Notably, SS-Voc requires no additional annotations or semantic knowledge; it simply shifts the burden from testing to training, leveraging the vocabulary to learn a better model. 
\end{definition}

The vocabulary $\mathcal{W}$ can come from a semantic embedding
space learned by word2vec~\cite{distributedword2vec2013NIPS} or
GloVec~\cite{GloVec}  on large-scale corpus; each vocabulary entity $w\in\mathcal{W}$ is represented
as a distributed semantic vector $\mathbf{u}\in\mathbb{R}^{d}$. 
Semantics of  embedding space help with knowledge transfer among
classes, and allow ZSL and open set image recognition. Note that such
semantic embedding spaces are equivalent to the ``semantic knowledge
base'' for ZSL defined in \cite{palatucci2009zero_shot} and hence make it 
appropriate to use SS-Voc in ZSL setting.

Assuming we can learn a mapping $g:\mathbb{R}^{p}\rightarrow\mathbb{R}^{d}$,
from image features to this semantic space, recognition can be carried
out using simple nearest neighbor distance, \eg, $f(\mathbf{x}^{*})=car$
if $g(\mathbf{x}^{*})$ is closer to $\mathbf{u}_{car}$ than to any
other word vector; $\mathbf{u}_{j}$ in this context can be interpreted
as the prototype of the class $j$. Thus the core question is then
how to learn the mapping $g(\mathbf{x})$ and what form of inference
is optimal in the semantic space. For learning we propose discriminative
maximum margin criterion that ensures that labeled samples $\mathbf{x}_{i}$
project closer to their corresponding class prototypes $\mathbf{u}_{z_{i}}$
than to any other prototype $\mathbf{u}_{i}$ in the open set vocabulary
$i\in\mathcal{W}\setminus z_{i}$.

\subsection{Learning Embedding}

Our maximum margin vocabulary-informed embedding learns the mapping
$g(\mathbf{x}):\mathbb{R}^{p}\rightarrow\mathbb{R}^{d}$, from low-level
features $\mathbf{x}$ to the semantic word space by utilizing maximum
margin strategy. Specifically, consider $g(\mathbf{x})=W^{T}\mathbf{x}$,
where%
\footnote{Generalizing to a kernel version is straightforward, see \cite{additiveKernel}.%
} $W\subseteq\mathbb{R}^{p\times d}$. Ideally we want to estimate
$W$ such that $\mathbf{u}_{z_{i}}=W^{T}\mathbf{x}_{i}$ for all labeled
instances in $\mathcal{D}_{s}$ (we would obviously want this to hold
for instances belonging to unobserved classes as well, but we cannot
enforce this explicitly in the optimization as we have no labeled
samples for them).

\vspace{0.05in}
\noindent
\textbf{Data Term:} The easiest way to enforce the above objective
is to minimize Euclidian distance between sample projections and appropriate
prototypes in the embedding space%
\footnote{Eq.(\ref{eq:data-embedding}) is also called data embedding \cite{unified_model} / compatibility function \cite{embedding_akata}.%
}: 
\begin{equation}
D\left(\mathbf{x}_{i},\mathbf{u}_{z_{i}}\right)=\parallel W^{T}\mathbf{x}_{i}-\mathbf{u}_{z_{i}}\parallel_{2}^{2}.\label{eq:data-embedding}
\end{equation}
\noindent We need to minimize this term with respect to each instance $\left(\mathbf{x}_{i},\mathbf{u}_{z_{i}}\right)$,
where $z_{i}$ is the class label of instance $\mathbf{x}_{i}$ in
$\mathcal{D}_{s}$. To prevent overfitting, we further regularize
the solution: 
\begin{equation}
\mathcal{L}\left(\mathbf{x}_{i},\mathbf{u}_{z_{i}}\right)=D\left(\mathbf{x}_{i},\mathbf{u}_{z_{i}}\right)+\lambda\parallel W\parallel_{F}^{2},\label{eq:least-square}
\end{equation}
where $\parallel\cdot\parallel_{F}$ indicates the Frobenius Norm.
Solution to the Eq.(\ref{eq:least-square}) can be obtained through
ridge regression. 

Nevertheless, to make the embedding more comparable to support vector regression (SVR),
we employ the maximal margin strategy -- $\epsilon-$insensitive smooth
SVR ($\epsilon-$SSVR) \cite{epsiSSVR} to replace the least square
term in Eq.(\ref{eq:least-square}). That is,

\begin{equation}
\mathcal{L}\left(\mathbf{x}_{i},\mathbf{u}_{z_{i}}\right)=\mathcal{L}_{\epsilon}\left(\mathbf{x}_{i},\mathbf{u}_{z_{i}}\right)+\lambda\parallel W\parallel_{F}^{2},\label{eq:SSVR}
\end{equation}

\noindent where $\mathcal{L}_{\epsilon}\left(\mathbf{x}_{i},\mathbf{u}_{z_{i}}\right)=\mathbf{1}^{T}\mid\xi\mid_{\epsilon}^{2}$;
$\lambda$ is regularization coefficient. $\left(|\xi|_{\epsilon}\right)_{j}=\mathrm{max}\left\{ 0,|W_{\star j}^{T}\mathbf{x}_{i}-\left(\mathbf{\mathbf{u}}_{z_{i}}\right)_{j}|-\epsilon\right\} $,
$|\xi|_{\epsilon}\in\mathbb{R}^{d}$, and $\left(\right)_{j}$ indicates
the $j$-th value of corresponding vector. $W_{\star j}$ is the $j$-th
column of $W$. The conventional $\epsilon-$SVR is formulated as
a constrained minimization problem, \ie, convex quadratic programming
problem, while $\epsilon-$SSVR employs quadratic smoothing \cite{Zhang:2004:SLS:1015330.1015332}
to make Eq.(\ref{eq:SSVR}) differentiable everywhere, and thus $\epsilon-$SSVR
can be solved as an unconstrained minimization problem directly%
\footnote{We found Eq.(\ref{eq:least-square}) and Eq.(\ref{eq:SSVR}) have
similar results, on average, but formulation in Eq.(\ref{eq:SSVR}) is more stable
and has lower variance. %
}. 

\vspace{0.05in}
\noindent
\textbf{Pairwise Term:} Data term above only ensures
that labelled samples project close to their correct prototypes. However,
since it is doing so for many samples and over a number of classes,
it is unlikely that all the data constraints can be satisfied exactly.
Specifically, consider the following case, if $\mathbf{u}_{z_{i}}$
is in the part of the semantic space where no other entities live
(\ie, distance from $\mathbf{u}_{z_{i}}$ to any other prototype
in the embedding space is large), then projecting $\mathbf{x}_{i}$
further away from $\mathbf{u}_{z_{i}}$ is asymptomatic, \ie, will
not result in misclassification. However, if the $\mathbf{u}_{z_{i}}$
is close to other prototypes then minor error in regression
may result in misclassification.

To embed this intuition into our learning, we enforce more
discriminative constraints in the learned semantic embedding space.
Specifically, the distance of $D\left(\mathbf{x}_{i},\mathbf{u}_{z_{i}}\right)$
should not only be as close as possible, but should also be smaller
than the distance $D\left(\mathbf{x}_{i},\mathbf{u}_{a}\right)$,
$\forall a\neq z_{i}$. Formally, we define the vocabulary pairwise
maximal margin term
\footnote{Crammer and Singer loss \cite{Tsochantaridis:2005:LMM:1046920.1088722,Crammer:2002:AIM:944790.944813}
is the upper bound of Eq (\ref{eq:vocab_maximal_margin}) and (\ref{eq:source_maximal_margin})
which we use to tolerate variants of $\mathbf{u}_{z_{i}}$ (e.g. 'pigs'
Vs. 'pig' in Fig. \ref{fig:AwA-openset_experiments}) and thus are
better for our tasks. 
}: 
\begin{equation}
\mathcal{M}_{V}\left(\mathbf{x}_{i},\mathbf{u}_{z_{i}}\right)=\frac{1}{2}\sum_{a=1}^{A_{V}}\left[C+\frac{1}{2}D\left(\mathbf{x}_{i},\mathbf{u}_{z_{i}}\right)-\frac{1}{2}D\left(\mathbf{x}_{i},\mathbf{u}_{a}\right)\right]_{+}^{2}\label{eq:vocab_maximal_margin}
\end{equation}
\noindent where $a\in\mathcal{W}_{t}$ is selected from the open vocabulary;
$C$ is the margin gap constant. Here, $\left[\cdot\right]_{+}^{2}$ indicates the
quadratically smooth hinge loss \cite{Zhang:2004:SLS:1015330.1015332}
which is convex and has the gradient at every point.  To speedup computation, we use the
closest $A_{V}$ target prototypes to each source/auxiliary prototype $\mathbf{u}_{z_{i}}$
in the semantic space. We also define similar constraints for the
source prototype pairs:
\begin{equation}
\mathcal{M}_{S}\left(\mathbf{x}_{i},\mathbf{u}_{z_{i}}\right)=\frac{1}{2}\sum_{b=1}^{B_{S}}\left[C+\frac{1}{2}D\left(\mathbf{x}_{i},\mathbf{u}_{z_{i}}\right)-\frac{1}{2}D\left(\mathbf{x}_{i},\mathbf{u}_{b}\right)\right]_{+}^{2}\label{eq:source_maximal_margin}
\end{equation}
\noindent where $b\in\mathcal{W}_{s}$ is selected from source/auxiliary
dataset vocabulary. This term enforces that $D\left(\mathbf{x}_{i},\mathbf{u}_{z_{i}}\right)$
should be smaller than the distance $D\left(\mathbf{x}_{i},\mathbf{u}_{b}\right)$,
$\forall b\neq z_{i}$. To facilitate the computation, we similarly
use closest $B_{S}$ prototypes that are closest to each prototype
$\mathbf{u}_{z_{i}}$ in the source classes. Our complete
pairwise maximum margin term is: 
\begin{equation}
\mathcal{M}\left(\mathbf{x}_{i},\mathbf{u}_{z_{i}}\right)=\mathcal{M}_{V}\left(\mathbf{x}_{i},\mathbf{u}_{z_{i}}\right)+\mathcal{M}_{S}\left(\mathbf{x}_{i},\mathbf{u}_{z_{i}}\right).\label{eq:maximal_margin_term}
\end{equation}

\noindent We note that the form 
of rank hinge loss in Eq.(\ref{eq:vocab_maximal_margin}) and 
Eq.(\ref{eq:source_maximal_margin}) is similar to DeViSE \cite{DeviseNIPS13}, but 
DeViSE only considers loss with respect to source/auxiliary data and prototypes.

\vspace{0.05in}
\noindent
\textbf{Vocabulary-informed Embedding:} The complete combined objective can now be written as: {
\vspace{-0.08in} 
\begin{eqnarray}
 W=\underset{W}{\mathrm{argmin}}\sum_{i=1}^{n_{T}}(\alpha\mathbb{\mathcal{L}_{\epsilon}}\left(\mathbf{x}_{i},\mathbf{u}_{y_{i}}\right)+ ~~~~~~~~~~~~~~~~~~~~~~~~~ \nonumber \\ (1-\alpha)\mathcal{M}\left(\mathbf{x}_{i},\mathbf{u}_{z_{i}}\right))+\lambda\parallel W\parallel_{F}^{2},\label{eq:formulation}
\end{eqnarray}
}

\noindent where $\alpha\in[0,1]$ is ratio coefficient of two terms. One practical advantage is that the objective
function in Eq.(\ref{eq:formulation}) is an unconstrained minimization
problem which is differentiable and can be solved with L-BFGS. 
$W$ is initialized with all zeros and converges in $10-20$ iterations.


\vspace{0.05in}
\noindent
\textbf{Fine-tuning Word Vector Space:} Above formulation works well
assuming semantic space is well laid out and linear mapping is sufficient.
However, we posit that word vector space itself is not necessarily
optimal for visual discrimination. Consider the following case: two
visually similar categories may appear far away in the semantic space.
In such a case, it would be difficult to learn a linear mapping that
matches instances with category prototypes properly. Inspired by this
intuition, which has also been expressed in natural language models
\cite{DBLP:journals/corr/BowmanPM14a}, we propose to fine-tune the
word vector representation for better visual discriminability.

One can potentially fine-tune the representation by optimizing
$\mathbf{u}_{i}$ directly, in an alternating optimization (\eg,
as in \cite{unified_model}). However, this is only possible for source/auxiliary
class prototypes and would break regularities in the semantic space,
reducing ability to transfer knowledge from source/auxilary to target
classes. Alternatively, we propose optimizing a global warping, $V$,
on the word vector space: 
\begin{eqnarray}
\left\{ W,V\right\}  =  \underset{W,V}{\mathrm{argmin}}\sum_{i=1}^{n_{T}}(\alpha\mathbb{\mathcal{L}_{\epsilon}}\left(\mathbf{x}_{i},\mathbf{u}_{y_{i}}V\right)+ ~~~~~~~~~~~~~~~~~~~  \nonumber\\
~~~~~~~ \left(1-\alpha\right)\mathcal{M}\left(\mathbf{x}_{i},\mathbf{u}_{z_{i}}V\right)) +\lambda\parallel W\parallel_{F}^{2}+\mu\parallel V\parallel_{F}^{2},  \label{eq:updating_prototypes}
\end{eqnarray}

\noindent where $\mu$ is regularization coefficient. Eq.(\ref{eq:updating_prototypes})
can still be solved using L-BFGS and $V$ is initialized using an
identity matrix. The algorithm first updates $W$ and then
$V$; typically the step of updating $V$ can converge within $10$
iterations and the corresponding class prototypes used for final
classification are updated to be $\mathbf{u}_{z_{i}}V$. 

\subsection{Maximum Margin Embedding Recognition}

\label{sec:recognition}

Once embedding model is learned, recognition in the semantic space
can be done in a variety of ways. We explore a simple alternative
to classify the testing instance $\mathbf{x}^{\star}$, 
\begin{equation}
z^{*}=\underset{i}{\mathrm{argmin}}\parallel W^T\mathbf{x}^{*}-\phi\left(\mathbf{u}_{i},V,W,\mathbf{x}^{*}\right)\parallel_{2}^{2}.\label{ZSL_classifier}
\end{equation}
\noindent Nearest Neighbor (NN) classifier directly measures the distance
between predicted semantic vectors with the prototypes in semantic
space, \ie,  $\phi\left(\mathbf{u}_{i},V,W,\mathbf{x}^{*}\right)=\mathbf{u}_{i}V$.
We further employ the k-nearest neighbors (KNN) of testing instances
to average the predictions, \ie, $\phi\left(\cdot\right)$ is averaging
the KNN instances of predicted semantic vectors.\footnote{This strategy is known as Rocchio algorithm
in information retrieval. Rocchio algorithm is a method for relevance feedback by
using more relevant instances to update the query instances for better
recall and possibly precision in vector space (Chap 14 in \cite{vector_space_classification}).
It was first suggested for use on ZSL in \cite{yanweiPAMIlatentattrib};
more sophisticated algorithms \cite{yanweiembedding,transferlearningNIPS}
are also possible.
} 



%
%

\section{Experiments}


\noindent \textbf{Datasets}. We conduct our experiments on Animals
with Attributes (AwA) dataset, and ImageNet $2012$/$2010$ dataset.
AwA consists of 50 classes of animals ($30,475$ images in total).
In \cite{lampert13AwAPAMI} standard split into 40 source/auxiliary
classes ($|\mathcal{W}_{s}|=40$) and 10 target/test classes ($|\mathcal{W}_{t}|=10$)
is introduced.
We follow this split for supervised and zero-shot learning. We use
OverFeat features (downloaded from \cite{semantic_graph}) on AwA to
make the results more easily comparable to state-of-the-art. ImageNet $2012$/$2010$
dataset is a large-scale dataset. We use $1000$ ($|\mathcal{W}_{s}|=1000$)
classes of ILSVRC $2012$ as the source/auxiliary classes and $360$
($|\mathcal{W}_{t}|=360$) classes of ILSVRC 2010 that
are not used in ILSVRC $2012$ as target data. We use pre-trained
VGG-19 model \cite{returnDevil2014BMVC} to extract deep features
for ImageNet. On both dataset, we use few instances from source
dataset to mimic human performance of learning from few examples and
ability to generalize.

\vspace{0.05in}
\noindent 
\textbf{Recognition tasks}. We consider three different settings in a variety of experiments 
(in each experiment we carefully denote which setting is used): 
\begin{description}
\vspace{-0.1in}
\item [] \textsc{Supervised} recognition, where learning is on source classes 
and we assume test instances come from same classes with $\mathcal{W}_{s}$
as recognition vocabulary; 
\vspace{-0.1in}
\item []  \textsc{Zero-shot} recognition, where learning is on source classes and 
we assume test instances coming from target dataset with $\mathcal{W}_{t}$
as recognition vocabulary; 
\vspace{-0.1in}
\item []  \textsc{Open-set} recognition, where we use entirely open vocabulary 
with $|\mathcal{W}|\approx310K$  and use 
test images from both source and target splits. 
\end{description}

\noindent \noindent \textbf{Competitors}. We compare the following models,
\begin{description}
\item [{SVM:}] SVM classifier trained directly on the training instances
of source data, without the use of semantic embedding. This is
the standard (\textsc{Supervised}) learning setting and the learned
classifier can only predict the labels in testing data of source classes. 
\vspace{-0.08in}
\item [{SVR-Map:}] SVR is used to learn $W$ and the recognition is done
in the resulting semantic manifold. This corresponds to only using
Eq.(\ref{eq:SSVR}) to learn $W$. 
\vspace{-0.08in}
\item [DeVise, ConSE, AMP:] 
To compare with state-of-the-art large-scale zero-shot learning approaches 
we implement DeViSE \cite{DeviseNIPS13} and ConSE \cite{ZSL_convex_optimization}\footnote{Code for \cite{DeviseNIPS13} and \cite{ZSL_convex_optimization} is not publicly available.}.
ConSE uses a multi-class logistic regression classifier for predicting class
probabilities of source instances; and the parameter T (number of
top-T nearest embeddings for a given instance) was selected from $\{1,10,100,1000\}$ that gives the best results. 
ConSE method in supervised setting works the same as SVR. We
use the AMP code provided  on the author webpage \cite{semantic_graph}. 
\vspace{-0.08in}

\item [{SS-Voc}:] We test three different variants of our method.  
\begin{description}
\vspace{-0.08in}
\item [{closed}] is a variant of our maximum margin leaning
of $W$ with the vocabulary-informed constraints only from known classes
(\ie, closed set $\mathcal{W}_{s}$). 
\vspace{-0.05in}
\item [{$W$}]  corresponds to our full model with maximum margin constraints
coming from both $\mathcal{W}_{s}$ and $\mathcal{W}_{t}$ (or $\mathcal{W}$).
We compute $W$ using Eq.(\ref{eq:formulation}), but without optimizing $V$.
\vspace{-0.05in}
\item [{full}] further fine-tunes the word vector space by also optimizing
$V$ using Eq.(\ref{eq:updating_prototypes}). 
\end{description}
\end{description}

\noindent 
\textbf{Open set vocabulary.} We use google word2vec to
learn the open set vocabulary set from a large text corpus of around
$7$ billion words: UMBC WebBase ($3$ billion words), the latest
Wikipedia articles ($3$ billion words) and other web documents ($1$
billion words). Some rare (low frequency) words and high frequency
stopping words were pruned in the vocabulary set: we remove words
with the frequency $<300$ or $>10\ million$ times. The result is a
vocabulary of around 310K words/phrases with $openness\approx1$, which
is defined as $openness=1-\sqrt{\left(2\times|\mathcal{W}_{s}|\right)/\left(|\mathcal{W}|\right)}$.
\cite{Scheirer_2013_TPAMI}.

\begin{table*}[!th]
\begin{centering}
{\small{}}%
\begin{tabular}{|l||c|c|c|c||c|c|c||ccc|}
\hline 
 & \multicolumn{3}{c|}{{\small{}Testing Classes}} &  &  &  &  & \multicolumn{3}{c|}{\textbf{\small{}SS-Voc}}\tabularnewline
 & {\small{}Aux } & {\small{}Targ. } & {\small{}Total } & {\small{}Vocab } & {\small{}Chance } & {\small{}SVM } & {\small{}SVR } & {\small{}closed } & {\small{}W } & {\small{}full }\tabularnewline
\hline 
\textsc{\small{}Supervised } & {\small{}\checkmark } &  & {\small{}40 } & {\small{}40 } & {\small{}$2.5$} & {\small{}$52.1$} & {\small{}$51.4/57.1$} & {\small{}$52.9/58.2$} & {\small{}$53.6/58.6$} & {\small{}$53.9/59.1$}\tabularnewline
\textsc{\small{}Zero-shot } &  & {\small{}\checkmark } & {\small{}10 } & {\small{}10 } & {\small{}$10$} & {\small{}- } & {\small{}$52.1/58.0$} & {\small{}$58.6/60.3$} & {\small{}$59.5/68.4$} & {\small{}$61.1/68.9$}\tabularnewline
\hline 
\end{tabular}
\par\end{centering}{\small \par}

{\tiny{}\protect\caption{{\small{}\label{tab:Standard-split-on}Classification accuracy
($\%$) on AwA dataset for }\textsc{\small{}Supervised}{\small{} and
}\textsc{\small{}Zero-shot}{\small{} settings for 100/1000-dim
word2vec representation. }}
}
\vspace{-0.1in}
\end{table*}

\noindent 
\textbf{Computational and parameters selection and scalability.}
All experiments are repeated $10$ times, to avoid
noise due to small training set size, and we report an average across
all runs. For all the experiments, the mean accuracy is reported,
\ie, the mean of the diagonal of the confusion matrix on the prediction
of testing data. We fix the parameters $\mu$ and $\lambda$ as $0.01$
and $\alpha=0.6$ in our experiments when only few training instances
are available for AwA (5 instances per class) and ImageNet (3 instances
per class). Varying values of $\lambda$, $\mu$ and $\alpha$
leads to $<1\%$ variances on AwA and $<0.2\%$ variances on ImageNet
dataset; but the experimental conclusions still hold. Cross-validation
is conducted when more training instances are available. $A_V$ and $B_S$ are set to $5$ to balance computational cost 
and efficiency of pairwise constraints.

To solve Eq.(\ref{eq:updating_prototypes}) at a scale, 
one can use Stochastic Gradient Descent (SGD) which makes
great progress initially, but often is slow when approaches a
solution. 
In contrast, the L-BFGS method mentioned above
can achieve steady convergence at the cost of computing the full objective
and gradient at each iteration. L-BFGS can usually achieve
better results than SGD with good initialization, however, is computationally 
expensive. 
To leverage benefits of both of these methods, 
we utilize a hybrid method to solve Eq.(\ref{eq:updating_prototypes})
in large-scale datasets:  the solver is initialized with few instances
to approximate the gradients using SGD first,
then gradually more instances are used and switch to L-BFGS is made 
with iterations. This solver is motivated by Friedlander \etal~\cite{FriedlanderSchmidt2012}, who theoretically
analyzed and proved the convergence for the hybrid optimization 
methods. 
In practice, we use L-BFGS and the Hybrid algorithms for AwA and ImageNet respectively. 
The hybrid algorithm can save between $20\sim50\%$ training time as compared
with L-BFGS.

\subsection{Experimental results on AwA dataset}


We report  AwA experimental results in Tab. \ref{tab:Standard-split-on},
which uses 100/1000-dimensional word2vec representation (\ie, $d=100/1000$).
We highlight the following observations: (1) \textbf{SS-Voc} variants 
have better classification accuracy than SVM and SVR. This validates
the effectiveness of our model. 
Particularly, the results of our \textbf{SS-Voc:full} are $1.8/2\%$ and $9/10.9\%$
higher than those of SVR/SVM on supervised and zero-shot recognition
respectively. Note that though the results of SVM/SVR are good for 
supervised recognition tasks (52.1 and 51.4/57.1 respectively),
we can further improve them, which we attribute to the more discriminative
classification boundary informed by the vocabulary.
(2) \textbf{SS-Voc:$W$} significantly, by up to $8.1\%$, improves  zero-shot recognition results of \textbf{SS-Voc:closed}. 
This validates the importance of information from open vocabulary. 
(3) \textbf{SS-Voc} benefits more from open set vocabulary as compared
to word vector space fine-tuneing. 
The results of supervised and zero-shot recognition of 
\textbf{SS-Voc:full} are $1/0.9\%$ and
$2.5/8.6\%$ higher than those of \textbf{SS-Voc:closed}. 

\vspace{0.05in}
\noindent 
\textbf{Comparing to state-of-the-art on ZSL:} We compare
our results with the state-of-the-art ZSL results on AwA dataset in
Tab.~\ref{tab:Zero-shot-Learning-Comparison}.  We compare \textbf{SS-Voc:full}
trained with all source instances, 800 (20 instances / class), and 200 instances
(5 instances / class). Our model achieves $78.3\%$ accuracy, which is remarkably 
higher than all previous methods. This is particularly impressive taking into 
account the fact that we use only a semantic space and no additional attribute
representations that many other competitor methods utilize.
Further, our  results with $800$ training instances, a small fraction of the 
$24,295$ instances used to train all other methods, already outperform all other approaches.
We argue that much of our success and improvement comes from a more discriminative
information obtained using an open set vocabulary and corresponding large margin constraints, 
rather than from the features, since our method improved $25.1\%$ as compared
with DAP \cite{lampert13AwAPAMI} which uses the same OverFeat features.
Note, our \textbf{SS-Voc:full} result is $4.4\%$ higher than the closest 
competitor \cite{embedding_akata}; this improvement is 
statistically significant. Comparing with our work, \cite{embedding_akata}
did not only use more powerful visual features (GoogLeNet Vs. OverFeat),
but also employed more semantic embeddings (attributes, GloVe%
\footnote{GloVe\cite{GloVec} can be taken as an improved version of word2vec.%
} and WordNet-derived similarity embeddings as compared to our word2vec). 

\begin{table}
\begin{centering}
\begin{tabular}{|c|c|c|c|}
\hline 
{}Methods  & S. Sp  & Features  & Acc. \tabularnewline
\hline 
\hline 
\textbf{SS-Voc:full} ~~~~~~~~~~~  & W  & $CNN_{\text{\tiny OverFeat}}$  & $\mathbf{78.3}$\tabularnewline
~~~~ $800$ instances  & W  & $CNN_{\text{\tiny OverFeat}}$  & $\mathbf{74.4}$\tabularnewline
~~~~ $200$ instances  & W  & $CNN_{\text{\tiny OverFeat}}$  & $68.9$ \tabularnewline
\hline 
\hline 
Akata \emph{et al.} \cite{embedding_akata}  & A+W  & $CNN_{\text{\tiny GoogleLeNet}}$  & $73.9$\tabularnewline
\hline 
TMV-BLP \cite{yanweiembedding}  & A+W  & $CNN_{\text{\tiny OverFeat}}$  & $69.9$\tabularnewline
\hline 
AMP (SR+SE) \cite{semantic_graph}  & A+W  & $CNN_{\text{\tiny OverFeat}}$  & $66.0$\tabularnewline
\hline 
DAP \cite{lampert13AwAPAMI}  & A  & $CNN_{\text{\tiny VGG19}}$  & $57.5$\tabularnewline
\hline 
PST\cite{transferlearningNIPS}  & A+W  & $CNN_{\text{\tiny OverFeat}}$ & $54.1$\tabularnewline
\hline 
DAP \cite{lampert13AwAPAMI}  & A  & $CNN_{\text{\tiny OverFeat}}$  & $53.2$\tabularnewline
\hline 
DS \cite{rohrbach2010semantic_transfer}  & W/A  & $CNN_{\text{\tiny OverFeat}}$  & $52.7$ \tabularnewline
\hline 
Jayaraman \emph{et al.} \cite{Jayaraman2014}  & A  & low-level  & $48.7$ \tabularnewline
\hline 
Yu \emph{et al.} \cite{Yucatergorylevel}  & A  & low-level  & $48.3$\tabularnewline
\hline 
IAP \cite{lampert13AwAPAMI}  & A  & $CNN_{\text{\tiny OverFeat}}$  & $44.5$\tabularnewline
\hline 
HEX \cite{Deng2014}  & A  & $CNN_{\text{\tiny DECAF}}$  & $44.2$\tabularnewline
\hline 
AHLE \cite{labelembeddingcvpr13}  & A  & low-level  & $43.5$\tabularnewline
\hline 
\end{tabular}
\par\end{centering}

\caption{\label{tab:Zero-shot-Learning-Comparison} {\bf Zero-shot 
comparison on AwA.} We compare the state-of-the-art ZSL results using different semantic
spaces (S. Sp) including word vector (W) and attribute (A). 
1000 dimension word2vec dictionary is used for SS-Voc.
(Chance-level
=$10\%$). Different types of CNN and hand-crafted low-level feature
are used by different methods. Except SS-Voc (200/800), all instances of source data
($24295$ images) are used for training. As a general reference, the
classification accuracy on ImageNet: $CNN_{\text{\tiny DECAF}}$
\textless{} $CNN_{\text{\tiny OverFeat}}$ \textless{} $CNN_{\text{\tiny VGG19}}$
\textless{} $CNN_{\text{\tiny GoogleLeNet}}$.}
\vspace{-0.15in}
\end{table}

\begin{figure}
\begin{centering}
\begin{tabular}{|l||c|c|c|c|}
\hline 
 & \multicolumn{3}{c|}{Testing Classes} & \tabularnewline
\multicolumn{1}{|r||}{\it AwA Dataset} & {Aux.} & {Targ.} & {Total } & { Vocab }\tabularnewline
\hline 
\textsc{\small Open-Set}$_{1K-NN}$     &          &           & 40~/~10   & $1000^{\star}$  \tabularnewline
\textsc{\small Open-Set}$_ {1K-RND}$  & (left) & (right) & 40~/~10  &  $1000^{\dagger}$ \tabularnewline
\textsc{\small Open-Set}$_{310K}$        &         &            & 40~/~10  &  $310K$      \tabularnewline
\hline 
\end{tabular} \\ 
\includegraphics[scale=0.25]{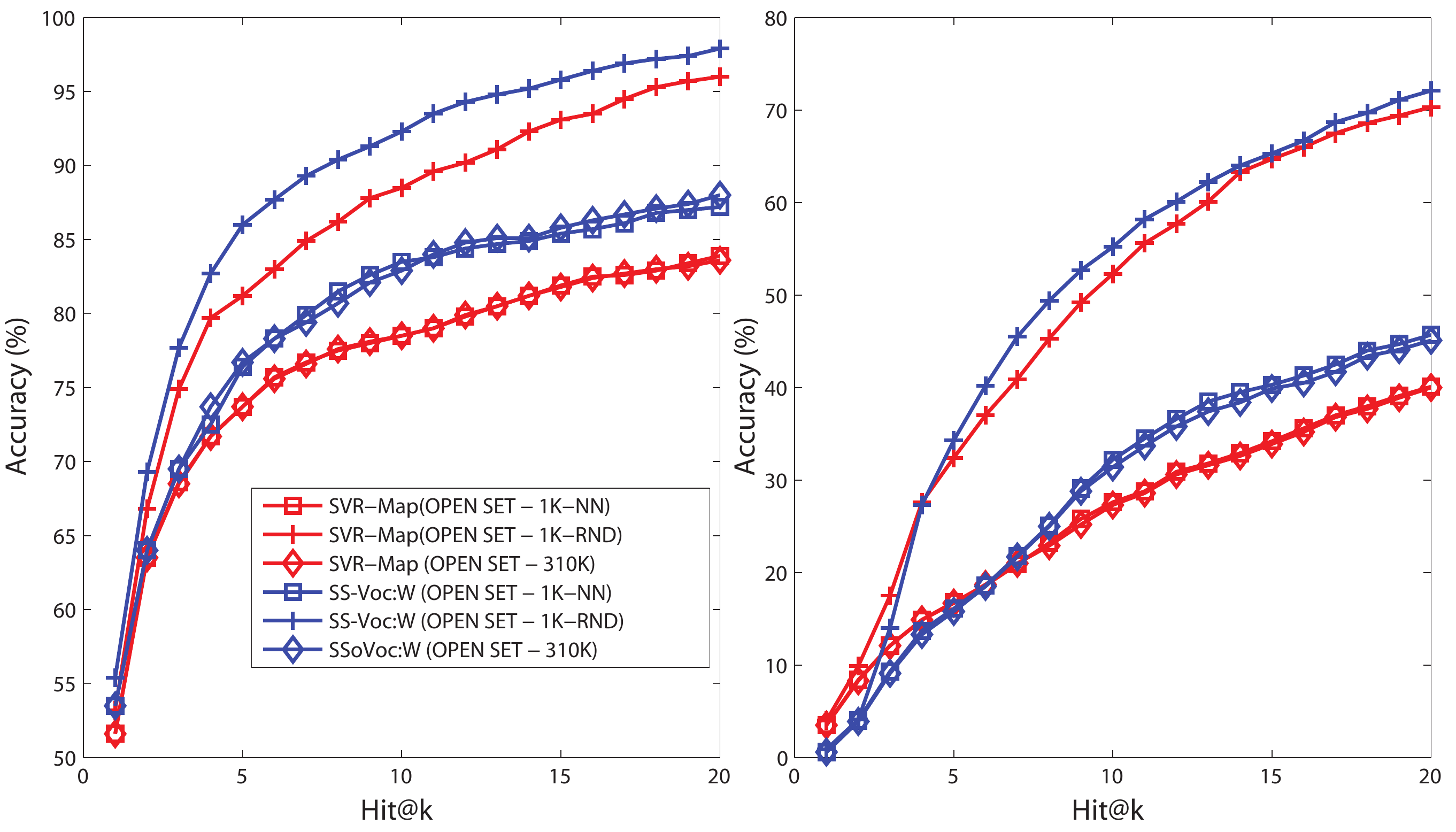}\tabularnewline

\par\end{centering}

\protect\protect\caption{\label{fig:AwA-openset_experiments} {\bf Open set recognition results
on AwA dataset:} Openness=$0.9839$. Chance=$3.2e-4\%$.
 Ground truth label is extended for its variants. For example,
we  count a correct label if a 'pig' image is labeled as 'pigs'. $\star$,$\dagger$:different $1000$ label settings. }
\vspace{-0.1in}
\end{figure}

\vspace{0.05in}
\noindent 
\textbf{Large-scale open set recognition:} Here we focus on \textsc{Open-set}$_{310K}$ setting with the large vocabulary of approximately 310K
entities; as such the chance performance of the task is much much
lower. In addition, to study the effect of performance as a function
of the open vocabulary set, we also conduct two additional experiments
with different label sets: (1)  \textsc{Open-set}$_{1K-NN}$: the 1000
labels from nearest neighbor set of ground-truth class prototypes
are selected from the complete dictionary of 310K labels. This corresponds
to an open set fine grained recognition; (2)  \textsc{Open-set}$_{1K-RND}$:
1000 label names randomly sampled from 310K set. The results are shown
in Fig. \ref{fig:AwA-openset_experiments}. Also note that we did
not fine-tune the word vector space (\ie, $V$ is an Identity matrix)
on \textsc{Open-set}$_{310K}$ setting since Eq~(\ref{eq:updating_prototypes})
can optimize a better visual discriminability only on a relative small
subset as compared with the 310K vocabulary.
While our \textsc{Open-set} variants do not assume that
test data comes from either source/auxiliary domain or target domain,
we split the two cases to mimic \textsc{Supervised} and \textsc{Zero-shot}
scenarios for easier analysis.

On \textsc{Supervised}-like setting, Fig.~\ref{fig:AwA-openset_experiments}
(left), our accuracy is better than that of SVR-Map on all the three
different label sets and at all hit rates. The better results are
largely due to the better embedding matrix $W$ learned by enforcing
maximum margins between training class name and open set vocabulary
on source training data. 

On \textsc{Zero shot}-like setting, our method still has a notable
advantage over that of SVR-Map method on Top-$k$ ($k>5$) accuracy,
again thanks to the better embedding $W$ learned by Eq.~(\ref{eq:formulation}).
However, we notice that our top-1 accuracy on \textsc{Zero shot}-like
setting is lower than SVR-Map method. We find that our method tends
to label some instances from target data with their nearest classes
from within source label set. For example, ``humpback whale''
from testing data is more likely to be labeled as ``blue whale''.
However, when considering Top-$k$ ($k>5$) accuracy, our method still
has advantages over baselines.

\vspace{0.02in}

\noindent 
\begin{figure*}
\begin{centering}
\includegraphics[scale=0.35]{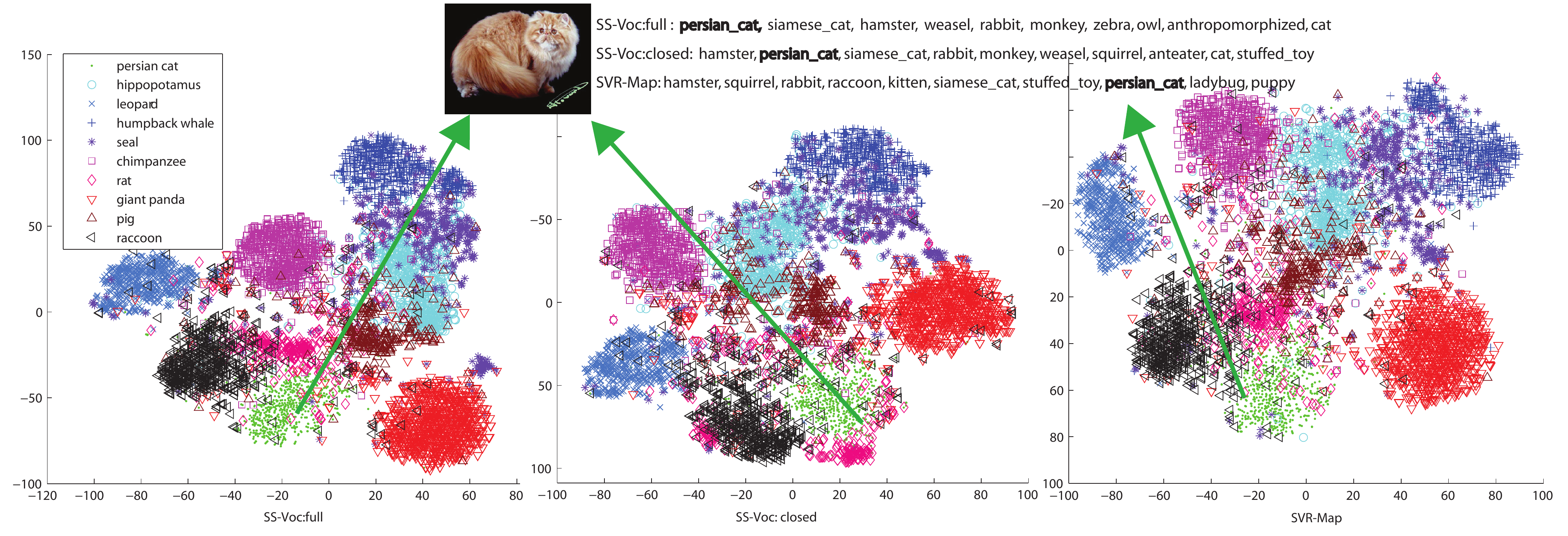}
\par\end{centering}

\protect\protect\caption{\label{fig:Qualitative-results.}t-SNE visualization of AwA 10 testing
classes. Please refer to Supplementary material for larger figure.}
\vspace{-0.15in}
\end{figure*}

\begin{table*}[!tbh]
\begin{centering}
{\small{}}%
\begin{tabular}{|l||c|c|c|c||c|c|c||ccc|}
\hline 
 & \multicolumn{3}{c|}{{\small{}Testing Classes}} &  &  &  & {\small{} } & \multicolumn{3}{c|}{\textbf{\small{}SS-Voc}}\tabularnewline
 & {\small{}Aux } & {\small{}Targ. } & {\small{}Total } & {\small{}Vocab } & {\small{}Chance } & {\small{}SVM } & {\small{}SVR } & {\small{}closed } & {\small{}W } & {\small{}full }\tabularnewline
\hline 
\textsc{\small{}Supervised } & {\small{}\checkmark } &  & {\small{}1000 } & {\small{}1000 } & {\small{}0.1} & {\small{}$33.8$} & {\small{}25.6 } & {\small{} 34.2} & {\small{}36.3 } & {\small{}37.1 }\tabularnewline
\textsc{\small{}Zero-shot } &  & {\small{}\checkmark } & {\small{}360 } & {\small{}360 } & {\small{}0.278 } & {\small{}- } & {\small{}4.1 } & {\small{} 8.0} & {\small{}8.2 } & {\small{}8.9}\tabularnewline
\hline 
\end{tabular}
\par\end{centering}{\small \par}

{\tiny{}\protect\protect\caption{\label{tab:ImageNet-2012/2010-dataset}The classification accuracy
($\%$) of ImageNet $2012$/$2010$ dataset on SUPERVISED and ZERO-SHOT
settings.}
}
\vspace{-0.15in}
\end{table*}

\vspace{-0.15in}

\subsection{Experimental results on ImageNet dataset}

\begin{figure}
\begin{centering}

\begin{tabular}{|l||c|c|c|c|}
\hline 
 & \multicolumn{3}{c|}{Testing Classes} &  \tabularnewline
\multicolumn{1}{|r||}{\it ImageNet Data} & Aux. & Targ.  & Total             & Vocab   \tabularnewline \hline
\textsc{Open-Set}$_{310K}$ & (left) & (right) & 1000~/~360 & $310K$ \tabularnewline
\hline 
\end{tabular}\tabularnewline

\includegraphics[scale=0.31]{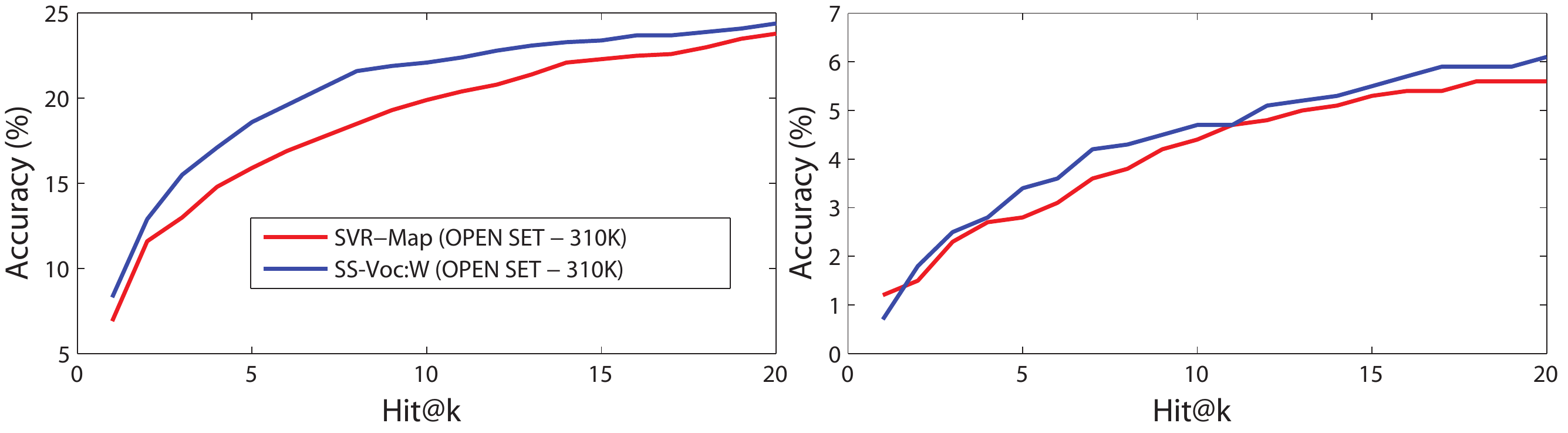}\tabularnewline

\par\end{centering}{\small \par}

\protect\protect\caption{\label{fig:ImageNet-2012/2010-open} {\bf Open set recognition results
on ImageNet 2012/2010 dataset:} Openness=$0.9839$. Chance=$3.2e-4\%$.
We use the synsets of each class--- a set of synonymous (word or prhase)
terms as the ground truth names for each instance. }
\vspace{-0.2in}
\end{figure}

We further validate our findings on large-scale ImageNet
2012/2010 dataset; 1000-dimensional word2vec representation is used here since
this dataset has larger number of classes than AwA. We highlight that
our results are still better than those of two baselines -- SVR-Map
and SVM on (\textsc{Supervised}) and (\textsc{Zero-shot}) settings
respectively as shown in Tab.~\ref{tab:ImageNet-2012/2010-dataset}.
The open set image recognition results are shown in Fig.~\ref{fig:ImageNet-2012/2010-open}.
On both \textsc{Supervised}-like and \textsc{Zero-shot}-like settings,
clearly our framework still has advantages over the baseline which
directly matches the nearest neighbors from the vocabulary by using
predicted semantic word vectors of each testing instance.

We note that \textsc{Supervised} SVM results ($34.61\%$) on ImageNet
are lower than $63.30\%$ reported in \cite{returnDevil2014BMVC},
despite using the same features. This is because only few, 3 samples
per class, are used to train our models to mimic human performance
of learning from few examples and illustrate ability of our model to learn
with little data. However, our semi-supervised vocabulary-informed
learning can improve the recognition accuracy on all settings. 
On open set image recognition, the performance has dropped
from $37.12\%$ (\textsc{Supervised}) and $8.92\%$ (\textsc{Zero-shot})
to around $9\%$ and $1\%$ respectively (Fig.~\ref{fig:ImageNet-2012/2010-open}). 
This drop is caused by the
intrinsic difficulty of the open set image recognition task ($\approx300\times$
increase in vocabulary) on a large-scale dataset. 
However, our performance is still better than the SVR-Map baseline
which in turn significantly better than the chance-level.

We also evaluated our model with larger number of training instances
($>3$ per class).
We observe that for standard supervised learning setting, the improvements 
achieved using vocabulary-informed learning tend to somewhat diminish as 
the number of training instances substantially grows. With large number
of training instances, the mapping between low-level image features
and semantic words, $g(\mathbf{x})$, becomes better behaved and effect of 
additional constraints, due to the open-vocabulary, becomes less 
pronounced. 

\vspace{0.1in}
\noindent
\textbf{Comparing to state-of-the-art on ZSL.} We compare our results
to several state-of-the-art large-scale zero-shot recognition models.
Our results, {\bf SS-Voc:full}, are better than those of ConSE, DeViSE
and AMP on both T-1 and T-5 metrics with a {\em very} significant margin 
(improvement over best competitor, ConSE, is 3.43 percentage points or 
nearly 62\% with $3,000$ training samples).  
Poor results of DeViSE with $3,000$ training instances
are largely due to the inefficient learning of visual-semantic
embedding matrix. AMP algorithm also relies on the embedding matrix
from DeViSE, which explains similar poor performance of AMP with $3,000$
training instances. 
In contrast, our {\bf SS-Voc:full} can leverage discriminative information from open
vocabulary and max-margin constraints, which helps improve performance.
For DeViSE with {\em all} ImageNet instances, we confirm the observation
in \cite{ZSL_convex_optimization} that results of ConSE are much better 
than those of DeViSE. Our results are a further significant improved from ConSE.

\subsection{Qualitative results of open set image recognition}

t-SNE visualization of AwA 10 target testing classes is shown in Fig.~\ref{fig:Qualitative-results.}.
We compare our \textbf{SS-Voc:full} with \textbf{SS-Voc:closed} and SVR. 
We note that (1) the distributions of 10 classes obtained using \textbf{SS-Voc} are more
centered and more separable than those of SVR (\eg, {\em rat}, {\em persian
cat} and {\em pig}), due to the data and pairwise maximum margin terms
that help improve the generalization of $g\left(\mathbf{x}\right)$
learned; (2) the distribution of different classes obtained using the full model
\textbf{SS-Voc:full} are also more separable than those of \textbf{SS-Voc:closed}, \eg, {\em rat}, {\em persian
cat} and {\em raccoon}. This can be attributed to the addition of the 
open-vocabulary-informed constraints during learning of $g\left(\mathbf{x}\right)$,
which further improves generalization.
For example, we show an open set recognition example image
of ``persian\_cat'', which is wrongly classified as a ``hamster'' 
by \textbf{SS-Voc:closed}.

Partial illustration of the embeddings learned for the ImageNet2012/2010 
dataset are illustrated in Figure~\ref{fig:intro}, where 4 source/auxiliary and 
2 target/zero-shot classes are shown. Again better separation among classes
is largely attributed to open-set max-margin constraints introduced in our 
\textbf{SS-Voc:full} model. Additional examples of miss-classified instances
are available in the supplemental material. 


\vspace{-0.10in}

\section{Conclusion and Future Work}
\vspace{-0.08in}
This paper introduces the problem of semi-supervised vocabulary-informed
learning, by utilizing open set semantic vocabulary to help train
better classifiers for observed and unobserved classes in supervised
learning, ZSL and open set image recognition settings. We formulate
semi-supervised vocabulary-informed learning in the maximum margin
framework. Extensive experimental results illustrate the efficacy
of such learning paradigm. Strikingly, it achieves competitive performance
with only few training instances and is relatively robust to large
open set vocabulary of up to $310,000$ class labels.

We rely on word2vec to transfer information
between observed and unobserved classes. In future, 
other linguistic or visual semantic embeddings could be explored instead, or 
in combination, as part of vocabulary-informed learning. 

\begin{table}[!t]
\begin{centering}
{\small{}}%
\begin{tabular}{|c|c|c|c|c|}
\hline 
{\small{}Methods } & {\small{}S. Sp} & {\small{}Feat.} & {\small{}T-1} & {\small{}T-5}\tabularnewline
\hline 
\hline 
\textbf{SS-Voc:full}{\small{} } & {\small{}W} & {\small{}$CNN_{\text{\tiny OverFeat}}$ } & {\small{}{\bf 8.9/9.5}} & {\small{}{\bf 14.9/16.8}}\tabularnewline
\hline 
{\small{}ConSE \cite{ZSL_convex_optimization}} & {\small{}W} & {\small{}$CNN_{\text{\tiny OverFeat}}$ } & {\small{}5.5/7.8} & {\small{}13.1/15.5}\tabularnewline
\hline 
{\small{}DeViSE \cite{DeviseNIPS13}} & {\small{}W} & {\small{}$CNN_{\text{\tiny OverFeat}}$ } & {\small{}3.7/5.2} & {\small{}11.8/12.8}\tabularnewline
\hline 
{\small{}AMP \cite{semantic_graph} } & {\small{}W} & {\small{}$CNN_{\text{\tiny OverFeat}}$ } & {\small{}3.5/6.1} & {\small{}10.5/13.1}\tabularnewline
\hline 
{\small{}Chance} & {\small{}--} & {\small{}--} & {\small{}$2.78e{\text -3}$} & {\small{}--}\tabularnewline
\hline 
\end{tabular}
\par\end{centering}{\small \par}

\protect\caption{{\bf ImageNet comparison to state-of-the-art on ZSL:} We compare the results
of using $3,000$/{\em all} training instances for all methods; T-1 (top 1) and T-5 (top 5) classification in \% is reported.}
\vspace{-0.20in}
\end{table}

{\small
\begin{spacing}{0.9}

 \bibliographystyle{abbrv}
\bibliography{egbib}
\end{spacing}
 }

\end{document}